\documentclass[runningheads]{llncs}
\usepackage[T1]{fontenc}
\usepackage{graphicx,verbatim}
\usepackage{multirow}%
\usepackage{amsmath,amssymb,amsfonts}%
\usepackage{dsfont}

\usepackage{amsthm}%
\usepackage{mathrsfs}%
\usepackage[title]{appendix}%
\usepackage{xcolor}%
\usepackage{textcomp}%
\usepackage{manyfoot}%
\usepackage{booktabs}%
\usepackage{algorithm}%
\usepackage{algorithmicx}%
\usepackage{algpseudocode}%
\usepackage{listings}%
\usepackage{array}%
\usepackage{hyperref}
\usepackage{soul} 
\usepackage{booktabs}  
\usepackage{array}     
\usepackage{textcomp}  
\usepackage{amssymb}   

\usepackage{color}

\begin{document}

\title{Multi-modal Representations for
Fine-grained Multi-label Critical View of Safety Recognition}
\titlerunning{CVS-AdaptNet}
%

\renewcommand{\thefootnote}{\fnsymbol{footnote}}
\footnotetext[1]{\textit{This manuscript has been accepted for publication and will appear in the proceedings of MICCAI 2025.}}

\author{Britty Baby\inst{1,4} \and
Vinkle Srivastav\inst{1,4} \and Pooja P. Jain \inst{1}\and Kun Yuan \inst{1,3}\and Pietro Mascagni \inst{2,4} \and Nicolas Padoy \inst{1,4}}

%
\institute{University of Strasbourg, CNRS, INSERM, ICube, UMR7357, Strasbourg, France 
\and
Fondazione Policlinico Universitario A. Gemelli IRCCS, Università Cattolica del Sacro Cuore, Rome, Italy \and
CAMP, Technische Universität München, Munich, Germany \and Institute of Image-Guided Surgery, IHU Strasbourg, Strasbourg, France \\ 
}
\authorrunning{B. Baby et al.}

\maketitle             
\begin{abstract}

The Critical View of Safety (CVS) is crucial for safe laparoscopic cholecystectomy, yet assessing CVS criteria remains a complex and challenging task, even for experts. Traditional models for CVS recognition depend on vision-only models learning with costly, labor-intensive spatial annotations. This study investigates how text can be harnessed as a powerful tool for both training and inference in multi-modal surgical foundation models to automate CVS recognition. Unlike many existing multi-modal models, which are primarily adapted for multi-class classification, CVS recognition requires a multi-label framework.  Zero-shot evaluation of existing multi-modal surgical models shows a significant performance gap for this task. To address this, we propose CVS-AdaptNet, a multi-label adaptation strategy that enhances fine-grained, binary classification across multiple labels by aligning image embeddings with textual descriptions of each CVS criterion using positive and negative prompts. By adapting PeskaVLP, a state-of-the-art surgical foundation model, on the Endoscapes-CVS201 dataset, CVS-AdaptNet achieves 57.6 mAP, improving over the ResNet50 image-only baseline (51.5 mAP) by 6 points. Our results show that CVS-AdaptNet’s multi-label, multi-modal framework, enhanced by textual prompts, boosts CVS recognition over image-only methods. We also propose text-specific inference methods, that helps in analysing the image-text alignment. While further work is needed to match state-of-the-art spatial annotation-based methods, this approach highlights the potential of adapting generalist models to specialized surgical tasks. Code: \url{https://github.com/CAMMA-public/CVS-AdaptNet}

\end{abstract}

\section{Introduction}

The development of models like CLIP~\cite{radford2021learning} has marked a significant advancement in multi-modal representation learning, enabling systems to interpret visual concepts through natural language. By aligning vision and language modalities on large-scale multi-modal datasets, the field of computer vision has progressed from task-specific approaches to more generalist models capable of handling multiple tasks with the same model~\cite{zou2023generalized,zou2024segment,ni2022expanding}. Recently, this trend has extended to medical and surgical AI, with models such as SurgVLP~\cite{yuan2025learning}, HecVL~\cite{yuan2024HecVL}, PeskaVLP~\cite{yuan2024procedure}, and VidLPRO~\cite{honarmand2024vidlpro} demonstrating promising generalization across datasets and tasks~\cite{twinanda2016endonet,nwoye2022rendezvous,lavanchy2024challenges}. While these models have shown success in coarse-grained tasks such as phase and tool recognition in zero-shot settings, their performance in more specialized surgical tasks remains under-explored.

One such specialized task is the assessment of the Critical View of Safety (CVS) in laparoscopic cholecystectomy—a crucial step in preventing bile duct injuries. Unlike traditional three-class classification, CVS assessment requires multi-label recognition, where an image can meet multiple criteria at the same time. This is particularly challenging in visually ambiguous images. Accurate CVS assessment requires a deep semantic understanding of hepatobiliary anatomy. However, even among experts, annotating CVS achievement is challenging. In the benchmark Endoscapes-CVS201 dataset~\cite{murali2023endoscapes}, annotations from three surgical experts yielded a low inter-annotator agreement (Cohen’s kappa = 0.38), highlighting the complexity of the task~\cite{mascagni2025endoscapes}.

Current state-of-the-art CVS assessment methods are vision-only and rely on pixel-wise spatial annotations to construct graph-based models that encode the anatomical relationships to predict CVS criteria~\cite{murali2023encoding,murali2023latent,satyanaik2024optimizing}.  However, the performance of these models is heavily dependent on the quality of anatomical segmentations. To investigate this dependency, we tested the SurgLatentGraph method~\cite{murali2023encoding} after removing a key structure (the cystic duct) from the annotations. This led to an 8-point drop in CVS mAP, emphasizing the critical role of detailed spatial annotations. Unfortunately, generating these fine-grained annotations is expensive, time-consuming, and prone to domain shift issues, reducing model generalizability across different surgical environments~\cite{satyanaik2024optimizing}.

Traditionally, specialized models have been developed for domain-specific surgical tasks. Given the rise of the multi-modal AI and the complexity of the CVS assessment task, in this work, we explore the reverse problem: \textit{Can current multi-modal surgical foundational models make use of their text-based knowledge to be adapted for a specialist surgical task?} To evaluate this, we tested PeskaVLP ~\cite{yuan2024procedure}, a state-of-the-art surgical foundational model, in a zero-shot setting on the CVS assessment task. The model achieved a mean Average Precision (mAP) of 26, an improvement of 7 points over the random baseline (mAP 19). While this demonstrates some capability, it also highlights the limitations of applying generalist multi-modal models directly to specialized tasks.

To address existing limitations, we propose CVS-AdaptNet, a novel adaptation strategy designed to utilize multi-modal foundation models for specialized multi-label surgical tasks. Our approach focuses on enhancing image-text alignment by incorporating naturally available textual descriptions of CVS criteria, without relying on spatial annotations such as bounding boxes or segmentation masks. Our key contributions include framing fine-grained CVS recognition as a multi-label, prompt-based task that captures the subjectivity of CVS assessment. Unlike prior CLIP adaptation and prompt tuning approaches, we tailor them to clinically detailed criteria instead of generic class names. We use diverse LLM-generated positive and negative prompts per criterion and apply KL divergence to model label ambiguity. Additionally, we propose multiple inference strategies to study the impact of prompt formulation and evaluate different multi-modal surgical foundation models for CVS classification.

Our experiments on Endoscapes-CVS201 dataset~\cite{murali2023endoscapes} show that well aligned image-text models can be successfully adapted to specific surgical tasks with improvements over image-only methods. While our results do not yet match pixel-wise segmentation methods, we believe this work lays the foundation for further exploration of multi-modal foundational models for specialized surgical tasks.

\section{Method}
\label{sec:method}
\begin{figure}[t]
\includegraphics[width=\textwidth]
{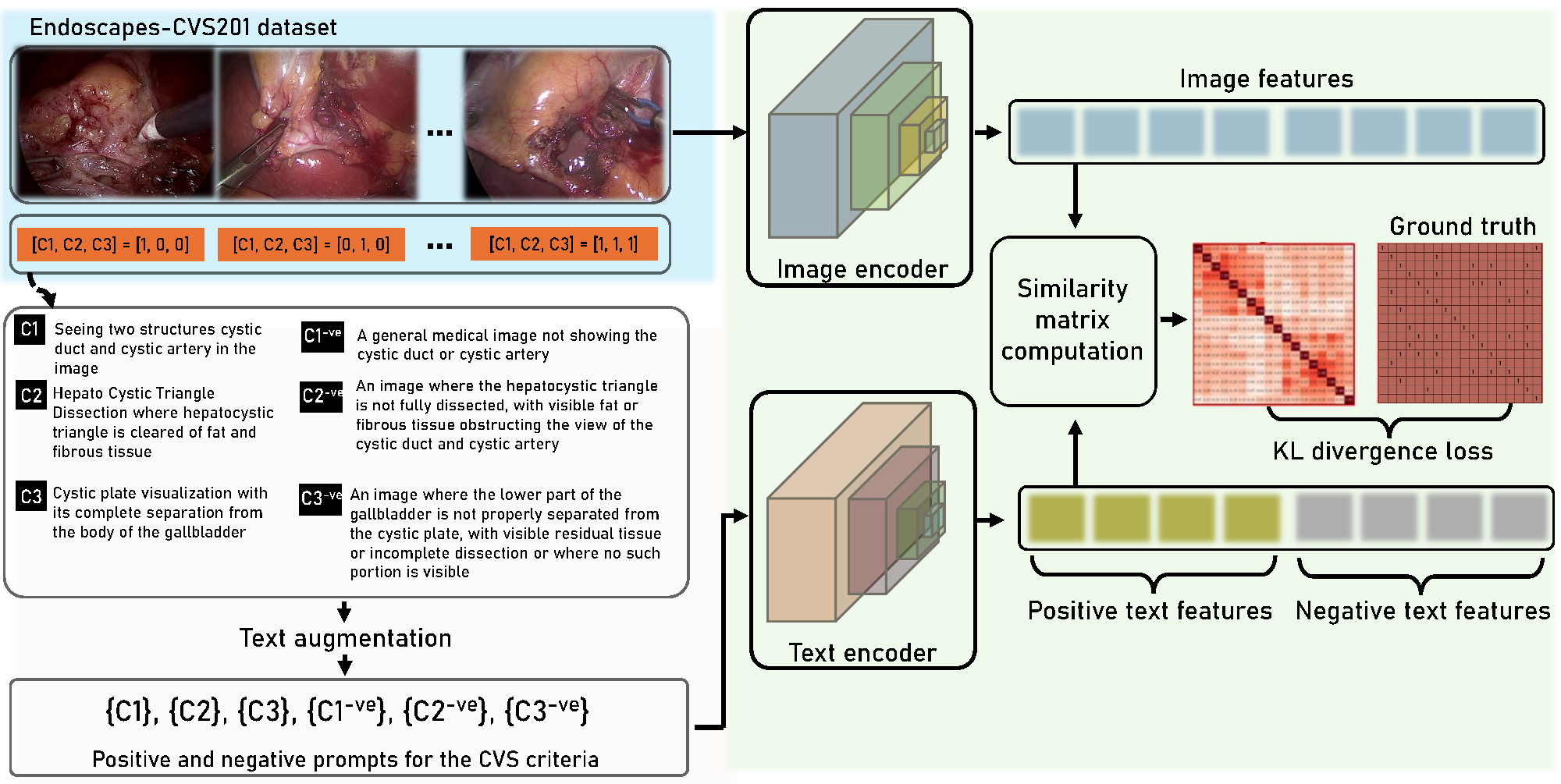}
\caption{Our proposed CVS-AdaptNet takes a batch of endoscopic images as input and uses an image encoder to extract visual features. A large language model (LLM) generates a diverse set of positive and negative prompts for each CVS criterion, which are used for text augmentation. During training, prompts indicating class presence or absence are randomly sampled and embedded using a text encoder. Visual and textual features are aligned using KL-divergence loss}
\label{fig:fig1}
\end{figure}

This section presents CVS-AdaptNet, our approach for adapting a multi-modal surgical foundation model to the task of Critical View of Safety (CVS) recognition. Unlike prior image-text adaptation strategies, such as those in Wang \textit{et al.}\cite{wang2023actionclip} for multi-class action recognition, we reformulate CVS recognition as a multi-label problem. Our method extends beyond binary classification by incorporating detailed text prompts—both positive and negative—for each CVS criterion. These prompts provide rich contextual cues, enabling the model to discern subtle distinctions in surgical images for fine-grained CVS assessment.

\noindent\textbf{Text Prompt Augmentation:} For each of the three CVS criteria defined by Strasberg \textit{et al.}\cite{strasberg1995analysis}: \\
    \textit{Criteria 1:}  ``the cystic duct and the cystic artery, connected to the gallbladder''\\
    \textit{Criteria 2:} ``a hepatocystic triangle cleared from fat and connective tissues''\\
    \textit{Criteria 3:} ``the lower part of the gallbladder separated from the liver bed'' \\
Based on this criteria, we generate - (a)\textit{Positive prompts}: Paraphrased descriptions indicating the presence of a criterion (e.g., ``Seeing two structures cystic duct and cystic artery in the image'' for Criterion 1). (b)\textit{Negative prompts}: Descriptions indicating absence (e.g., ``A general medical image not showing cystic duct or cystic artery'' for Criterion 1). To enhance prompt variability, we employ a large language model (LLM) to augment texts by generating diverse paraphrases, ensuring robustness~\cite{chatgpt2024}. This process is illustrated in Figure~\ref{fig:fig1}.

\noindent\textbf{Training:} Let \( x_j \) denote the \( j \)-th surgical image in a batch, and \({Y} = \{y_1, y_2, .., y_K\} \) represent the set of binary labels for the $K$ CVS criteria (\( K = 3 \)). Each \( y_{ji} \in \{0, 1\} \) indicates the absence (0) or presence (1) of criterion \( i \) corresponding to the j-th image. The image encoder \( g_I \) and text encoder \( g_W \) produce feature representations:
\begin{equation}
    v_j = g_I(x_j), \quad t_{ji}^+ = g_W(y_{ji}^+),  \quad t_{ji}^- = g_W(y_{ji}^-)
\end{equation}
where \( v_j \) is the image feature, and \( t_{ji}^+ \) and \( t_{ji}^- \) are the text features for positive and negative prompts of criterion \( i \), respectively. During training, for each image-criterion pair, we randomly select a positive prompt from the paraphrased set if \( y_{ji} = 1 \), or a negative prompt if \( y_{ji} = 0 \). For each criterion \( i \), we define the text representation:
\begin{equation}
t_{ji} = \mathds{1}_{\{y_{ji} = 1\}} \cdot t_{ji}^+ + \mathds{1}_{\{y_{ji} = 0\}} \cdot t_{ji}^-
\label{eq:label_selector}
\end{equation}
where \( \mathds{1} \) is the indicator function selecting the appropriate prompt feature based on the label.
We compute cosine similarity between image and text features:
\begin{equation}
\label{eq:similarity}
\begin{aligned}
s(v_j, t_{ji}) = \frac{v_j \cdot t_{ji}}{\|v_j\| \|t_{ji}\|}, \quad s(t_{ji}, v_j) = \frac{t_{ji} \cdot v_j}{\|t_{ji}\| \|v_j\|} 
\end{aligned}
\end{equation}
where \( v_j \) is the encoded feature of image \( x_j \), and \( t_{ji} \) is the text feature for criterion \( i \) corresponding to the j-th image.
For a given criterion \( i \), let's assume the batch contains a mix of prompts matching \(x_j\) and not matching \(x_j\), we obtain normalized similarity scores, over all these text prompts:

\begin{equation}
\label{eq:softmax}
p_{x_{j} \to
y_{ji}}(x_{j}) = \frac{\exp(s(v_{j}, t_{ji}) / \tau)}{\sum_{k=1}^{N} \exp(s(v_{j}, t_{ki}) / \tau)},  \quad   
p_{y_{ji} \to x_{j}}(y_{ji}) = \frac{\exp(s(t_{ji}, v_{j}) / \tau)}{\sum_{k=1}^{N} \exp(s(t_{ki}, v_{j}) / \tau)}
\end{equation}
where  \(s(v_{j}, t_{ji} )\) is the similarity score between the input image feature \(v_{j}\), to the corresponding text feature \(t_{ji}\), \( \tau \) is a learnable temperature parameter that controls the sharpness of the distribution, \( N \) is the batch size. The denominator sums over all the N text prompts in the batch. 
This normalization ensures that the similarity scores are converted into probability distributions, facilitating a structured contrastive learning process.

To model the inherent annotator ambiguity and variability in CVS labels, we use Kullback-Leibler (KL) divergence as a contrastive loss. Unlike Binary Cross-Entropy, which assumes independent binary labels, and InfoNCE, which enforces a rigid 1:N contrast, KL divergence accommodates many-to-many matches across prompts and images. For each image \( x_j \) and criterion \( i \), the loss is:
\begin{equation}
\label{eq:kl_divergence}
L_i(x_j) = \frac{1}{2} \left[ \mathcal{KL}(p_{x_j \to y_{ji}}(x_j) \,\|\, q_{x_j \to y_{ji}}(x_j)) + \mathcal{KL}(p_{y_{ji} \to x_j}(y_{ji}) \,\|\, q_{y_{ji} \to x_j}(y_{ji})) \right]
\end{equation}
where \( p_{x_j \to y_{ji}}(x_j) \) is the predicted probability for image-to-text alignment, and \( q_{x_j \to y_{ji}}(x_j) \in \{0, 1\} \) is the ground truth (1 for a positive match, 0 otherwise). Similarly, \( p_{y_{ji} \to x_j}(y_{ji}) \) and \( q_{y_{ji} \to x_j}(y_{ji}) \) align text-to-image predictions. KL divergence encourages the model to learn a sharper, more discriminative separation between matching and non-matching pairs, as it uses the full batch for contrastive learning.
The total loss over a batch \( B \) is:
\begin{equation}
    L = \sum_{i=1}^{K} \mathbb{E}_{(x_j, y_{ji}) \sim B} \left[ L_i(x_j) \right]
\end{equation}

\noindent where \( \mathbb{E} \) denotes the expectation (average) over the batch \( B \) and the \( K \) criteria, ensuring balanced optimization.

\noindent\textbf{Inference Strategies:} During inference, we explore three strategies to enhance robustness and adaptability in real-world surgical settings. Unlike traditional methods bound by fixed output vectors, image-text models offer the flexibility to query visual and textual information in various ways:\\
\textbf{1. Standard Inference}: For a test image \( x \), we extract its feature \( v = g_I(x) \) and compute its cosine similarity to a fixed set of clinician-selected text features \( T = \{t_1, t_2, t_3\} \), one per criterion (distinct from training prompts):
\begin{equation}
s(v, t_i) = \frac{v \cdot t_i}{\|v\| \|t_i\|}.
\label{eq:standard_inference}
\end{equation}
Class probabilities are then obtained by applying the sigmoid function to these similarities, denoted \( \hat{y}_i = \sigma(s(v, t_i)) \), where \( \sigma(z) = 1 / (1 + e^{-z}) \). This method is simple but assumes independence between criteria and may struggle with ambiguous cases. Additionally, using sigmoid over logits introduces uncertainty in selecting an appropriate decision threshold.\\
\textbf{2. Positive-Negative Inference:} To evaluate further, we adopt a contrastive approach by defining two sets of text embeddings for each criterion: positive prompts \( T^+ = \{t^+_1, t^+_2, t^+_3\} \) and negative prompts \( T^- = \{t^-_1, t^-_2, t^-_3\} \). These sets explicitly describe the presence or absence of key surgical features. For an image feature \( v = g_I(x) \), we compute cosine similarities for each criterion \( i \):
\begin{equation}
s^+_i = \frac{v \cdot t^+_i}{\|v\| \|t^+_i\|}, \quad s^-_i = \frac{v \cdot t^-_i}{\|v\| \|t^-_i\|}.
\end{equation}
For each criterion, we stack the positive and negative similarities \( [s^+_i, s^-_i] \) and apply the softmax function to normalize them into probabilities:
\begin{equation}
p^+_i = \frac{e^{s^+_i}}{e^{s^+_i} + e^{s^-_i}}, \quad p^-_i = \frac{e^{s^-_i}}{e^{s^+_i} + e^{s^-_i}}.
\end{equation}
The final probability for the positive class is taken as \( \hat{y}_i = p^+_i \). This tests contrasting ability to select the positive prompt.\\
\textbf{3. Multi-Class Inference}: To capture nuanced combinations of surgical features, we expand the text embeddings to a richer set \( T' = \{t'_1, t'_2, \ldots, t'_8\} \), where each \( t'_i \) describes a unique scenario by combining aspects of the original criteria \( t_1, t_2, t_3 \). For an image feature \( v = g_I(x) \), we compute similarities:
\begin{equation}
s(v, t'_i) = \frac{v \cdot t'_i}{\|v\| \|t'_i\|}.
\end{equation}
These similarities are normalized across all eight descriptions using softmax:
\begin{equation}
p_i = \frac{e^{s(v, t'_i)}}{\sum_{j=1}^8 e^{s(v, t'_j)}}.
\end{equation}
To align with the original three criteria, we aggregate probabilities by mapping each \( t'_i \) to the criteria it includes (e.g., \( \hat{y}_1 = \sum_{t'_i \text{ includes } t_1} p_i \)). This approach evaluates robustness by allowing the model to match images to a broader range of descriptions, accommodating complex surgical scenes.

\section{Datasets and Experiments}
\label{sec:datasets}
We evaluate HecVL~\cite{yuan2024HecVL}, SurgVLP~\cite{yuan2025learning}, and PeskaVLP~\cite{yuan2024procedure} as surgical multi-modal foundational models, pre-trained on surgical lecture videos from the SVL dataset~\cite{yuan2025learning}.

For evaluation, we use the Endoscapes-CVS201 dataset~\cite{murali2023endoscapes}, containing 11,090 frames (6960 train, 2331 val, 1799 test) annotated with three CVS criteria per frame. We do not use bounding boxes or segmentation masks for training. As a lower baseline, we compare our models against an ImageNet-pretrained ResNet50~\cite{he2016deep}, which also uses only CVS labels without any spatial annotations. Further, we compare with image-only classifiers initialized with different pre-training strategies.

\begin{table}[t]
\centering
\caption{Comparison on EndoScapes-CVS201: Results (mean±std) are averaged over 3 runs; results without std are from references. Average Precision (AP) is reported per criterion and as mean AP (mAP); * denotes use of graphs/spatial annotations.}
\label{table:Finetuning_results}
\resizebox{0.7\textwidth}{!}{%
\begin{tabular}{lcccccc}
\toprule
\textbf{Architecture} & \textbf{Initialization} & \textbf{mAP} & \textbf{C1} & \textbf{C2} & \textbf{C3} \\
\midrule
\multicolumn{6}{c}{\textbf{Image-only}} \\
\midrule
ResNet50 \cite{murali2023endoscapes} & ImageNet & 51.5 & 42.9 & 46.5 & 65.1 \\
ResNet50-MoCov2 \cite{murali2023endoscapes} & SSL\_pretrained & \underline{57.4} & 46.4 & \underline{\textbf{56.5}} & \underline{\textbf{69.4}} \\
\midrule
SurgVLP-vision \cite{yuan2025learning} & SurgVLP & 51.4±1.5 & \underline{51.6±1.5} & 41.1±2.4 & 61.4±3.6 \\
HecVL-vision \cite{yuan2024HecVL} & HecVL & 50.0±0.5 & 47.8±1.2 & 42.4±1.2 & 59.7±2.7 \\
PeskaVLP-vision \cite{yuan2024procedure} & PeskaVLP & 48.9±2.3 & 48.3±3.9 & 42.4±4.4 & 56.2±1.5 \\
\midrule
\multicolumn{6}{c}{\textbf{Image+Text}} \\
\midrule
\textbf{Standard} \\
CVS-AdaptNet & ResNet50 + BioClinicalBERT & 50.3±1.6 & 50.2±3.0 & 44.9±1.7 & 55.7±0.7 \\
CVS-AdaptNet & SurgVLP \cite{yuan2025learning} & 51.4±0.5 & 53.5±0.1 & 43.9±1.7 & 56.6±0.4 \\
CVS-AdaptNet & HecVL \cite{yuan2024HecVL} & 51.7±0.4 & 51.2±0.1 & 44.6±0.7 & 59.2±1.3 \\
CVS-AdaptNet & PeskaVLP \cite{yuan2024procedure} & \underline{\textbf{57.6±0.2}} & \underline{\textbf{54.5±0.7}} & \underline{55.9±0.3} & \underline{62.4±0.5} \\
\midrule
\textbf{Positive-Negative} \\
CVS-AdaptNet & ResNet50 + BioClinicalBERT & 50.0±1.8 & 51.0±3.0 & 42.5±1.6 & 56.4±1.1\\
CVS-AdaptNet & SurgVLP \cite{yuan2025learning} & 50.4±0.9 & 53.6±0.4 & 42.5±2.3 & 55.2±1.0 \\
CVS-AdaptNet & HecVL \cite{yuan2024HecVL} & 51.3±0.9 & 49.0±1.4 & 45.4±0.8 & 59.7±1.1 \\
CVS-AdaptNet & PeskaVLP \cite{yuan2024procedure} & \underline{56.8±0.4} & \underline{53.7±0.2} & \underline{54.6±0.9} & \underline{62.0±0.7} \\
\midrule
\textbf{Multi-class} \\
CVS-AdaptNet & ResNet50 + BioClinicalBERT & 46.8±2.5 & 44.6±2.4 & 45.6±1.5 & 50.0±6.6 \\
CVS-AdaptNet & SurgVLP \cite{yuan2025learning} & 35.9±0.1 & 39.2±1.6 & 35.2±0.8 & 33.1±0.8 \\
CVS-AdaptNet & HecVL \cite{yuan2024HecVL} & 52.0±0.9 & 53.8±1.2 & 44.4±1.0 & 57.8±2.5 \\
CVS-AdaptNet & PeskaVLP \cite{yuan2024procedure} & \underline{57.6±0.3} & \underline{54.1±1.0} & \underline{55.4±0.4} & \underline{63.3±0.1} \\
\midrule
\multicolumn{6}{c}{\textbf{Image + Segmentation Masks}} \\
\midrule
LG-CVS* \cite{murali2023latent} & Mask-RCNN* & 67.3 & 69.5 & 60.7 & 71.8 \\
\bottomrule
\end{tabular}%
}
\end{table}
\noindent \textbf{Zero-shot:} We first evaluate the pre-trained models~\cite{yuan2025learning,yuan2024HecVL,yuan2024procedure} on the Endoscapes-CVS201 test set using standard inference (eq.~\ref{eq:standard_inference}).

\noindent \textbf{Fully supervised fine-tuning:} We fine-tune models on Endoscapes-CVS201 in a supervised manner. Image-only methods use one-hot labels for multi-label classification, while image-text methods initialize from pre-trained models and adapt using CVS-AdaptNet. Confidence scores from three annotators are rounded to 0 or 1 for fair comparison with baselines~\cite{murali2023encoding}.

\noindent \textbf{Implementation:} We use ResNet50~\cite{he2016deep} as the vision encoder and BioClinicalBERT~\cite{huang2019clinicalbert} as the text encoder, adopting pre-trained weights from~\cite{yuan2025learning,yuan2024HecVL,yuan2024procedure}. We used the image size of $360$x$640$ and a center crop of $224$. We train our model using the Adam optimizer with an initial rate of $1\mathrm{e}{-5}$, decayed by the cosine annealing rule. We apply RandAugment \cite{cubuk2020randaugment} for images and use proposed text augmentations for training. The temperature parameter $\tau$ is learnable (initialized at $2.6593$), with a learning rate of $0.001$. We train both the vision and text encoders together for 20 epochs with a batch size of $64$ using a single RTX A5500 GPU ($\approx 4 \text{ hours}$). The training was performed using the PyTorch framework, and the best model is selected based on validation CVS mAP. 

\noindent \textbf{Evaluation Metrics:} We compute average precision per criterion (C1, C2, C3) and mean average precision (mAP) at the frame level, using similarity scores between test images and text prompts as described in section ~\ref{sec:method}.

\section{Results and Discussions}
\label{sec:results}

\textbf{Zero-shot inference:} We evaluate the zero-shot performance of SurgVLP~\cite{yuan2025learning}, HecVL~\cite{yuan2024HecVL}, and PeskaVLP~\cite{yuan2024procedure}, obtaining mAP scores of 26.18, 28.46, and 26.64, respectively. While these models outperform a random baseline (mAP: 19), they are not yet suitable for direct application.\\
\textbf{Fully supervised fine-tuning:} We compare different setups (shown in Table~\ref{table:Finetuning_results})\\
\textbf{Image-Only:} Baselines trained using one-hot labels.
\textit{ResNet50:} A standard ResNet50 classifier with a modified final layer to output three CVS scores \cite{murali2023latent}. We then use the same setup with different initializations:  
\textit{ResNet50-MoCov2:} Initialized with a MoCov2-pretrained backbone, from a self-supervised learning pipeline on surgical videos \cite{murali2023latent}.  
\textit{SurgVLP-vision:} Initialized with a pretrained backbone from \cite{yuan2025learning}.  
\textit{HecVL-vision:} Initialized with a pretrained backbone from \cite{yuan2024HecVL}.  
\textit{PeskaVLP-vision:} Initialized with a pretrained backbone from \cite{yuan2024procedure}.\\
\textbf{Image+Text:}  
We evaluate CVS-AdaptNet with different pre-training initializations. ResNet50 + BioClinicalBERT (without surgery-specific pre-training) performs slightly worse than image-only models, indicating weak vision-text alignment. Minimal improvement is seen with SurgVLP~\cite{yuan2025learning} and HecVL~\cite{yuan2024HecVL}, suggesting weak alignment. However, PeskaVLP achieves a significant boost (mAP: 57.6 ± 0.2), showing superior image-text alignment. Interestingly, ResNet50-MoCov2 performs comparably, suggesting that both image-only and multi-modal pipelines can yield similar outcomes, and improved pre-training architectures could enhance results further. Our different inference strategies show that adaptation using PeskaVLP consistently outperforms other foundation models, highlighting its discriminative ability to adapt to varying text inputs.

\noindent \textbf{Image+Segmentation Masks:} 
We compare the upper baseline reported on Endoscapes-CVS201 using the SurgLatentGraph \cite{murali2023latent}. This model uses two-stage learning and extra annotations. First the segmentation masks are learned and then relation graph of the anatomical structures giving superior performance.

\noindent \textbf{Ablations:}
In our ablation studies, we investigate how different data processing techniques and prompt designs affect performance. We compare two prompting strategies: Fixed-class prompting, which employs only fixed prompts for each image as defined in the original work \cite{strasberg1995analysis}; and Text-augmented prompting, which incorporates both positive and negative prompts generated from LLM~\cite{chatgpt2024}. Additionally, we examine how the level of detail in the prompts influences results, with findings summarized in Table~\ref{tab:ablation_study}. It is important to note that, \textit{Random} text inputs tend to confuse the vision encoder and degrade performance, whereas carefully crafted prompts enhances prediction accuracy. With fixed-class prompting, the model tends to overfit to the input texts and fails to perform on our other inference strategies; positive-negative inference and multi-class inference. 

\begin{table}[t]
    \centering
    \caption{Performance of different text types in CVS recognition task using standard inference. \checkmark shows text-augmented prompting using positive and negative prompts.}
    \resizebox{1.0\textwidth}{!}{
    \begin{tabular}{l|c|c|cccc|c}
        \toprule
        \textbf{Text Type} & \textbf{Initialization} & \textbf{Aug} & \textbf{mAP} & \textbf{C1} & \textbf{C2} & \textbf{C3} & \textbf{Observations} \\
        \midrule
        Random  & ResNet50+BioClinicalBERT & \checkmark & 45.09 & 46.73 & 33.74 & 54.81 & Worst - misaligned text hurts. \\
        Generic Surgical  & ResNet50+BioClinicalBERT & \checkmark & 49.31 & 50.01 & 46.83 & 51.11 & Slight improvement, but still weak. \\
        Fixed Class Prompts & PeskaVLP \cite{yuan2024procedure} & $\times$  & 54.26 & 51.26 & 53.95 & 57.55 & Class Text + Pretraining helps  \\
        Detailed-Anatomical  & PeskaVLP \cite{yuan2024procedure} & \checkmark & 53.20 & 49.23 & 53.11 & 57.26 & Fine details don’t always help \\
        Medium Detailed   &
        PeskaVLP \cite{yuan2024procedure} & \checkmark & \textbf{57.54} & \textbf{53.78} & \textbf{56.3} & \textbf{62.54} & \textbf{Best - discriminative text + Pretraining helps}  \\
        \bottomrule
    \end{tabular}}
    \label{tab:ablation_study}
\end{table}

\noindent \textbf{Limitations and Method Scope:}
We acknowledge the limitation of using a single dataset, as few CVS datasets exist. However, our method is designed to generalize to descriptive, multi-label surgical recognition tasks, where dense annotations are costly. We excluded adaptation methods such as DualCoOP~\cite{sun2022dualcoop} and MMA~\cite{yang2024mma} due to architectural incompatibilities with pre-trained surgical foundation models (SurgVLP~\cite{yuan2025learning}, HecVL~\cite{yuan2024HecVL}, PeskaVLP~\cite{yuan2024procedure}) that hinder direct adaptation.

\section{Conclusion}
\label{sec:conclusion}

Recent advances in multi-modal pre-training show that generic visual and textual representations can support a wide range of downstream tasks without explicit annotations. Building on this, we explore how natural language descriptions enhance performance in specialized surgical applications. Our method adapts to fine-grained, multi-label CVS recognition by using text prompts and CVS criteria descriptions to align visual and textual features without relying on extra annotations. The results highlight the potential of multimodal models to improve adaptability in real-world surgical settings and enhance patient safety.

\begin{credits}
\subsubsection{\ackname}

This work has received funding from the European Union (ERC, CompSURG, 101088553). Views and opinions expressed are however those of the authors only and do not necessarily reflect those of the European Union or the European Research Council. Neither the European Union nor the granting authority can be held responsible for them. 
Partial support came from the French ANR (Grant ANR-10-IAHU-02). The authors thank the University of Strasbourg’s High Performance Computing Center for scientific support and resource access, partially funded by Equipex Equip@Meso (Investissements d’Avenir) and CPER Alsacalcul/Big Data.

\subsubsection{\discintname}
The authors have no competing interests to declare relevant to this article.

\end{credits}
%
%

\end{document}